\documentclass[letterpaper, 10 pt, conference]{ieeeconf}  

\IEEEoverridecommandlockouts                              

\usepackage{graphicx, wrapfig}
\usepackage[format=plain,
            labelfont=it,
            labelsep=period]{caption}
\usepackage{subcaption}
\usepackage[colorlinks,urlcolor=blue,linkcolor=blue]{hyperref}
\usepackage{fontawesome} 
\newcommand{\extweblink}{{\footnotesize{\faExternalLink}}}
\newcommand{\webpage}{\href{https://sites.google.com/view/modem-v2}{sites.google.com/view/modem-v2\extweblink}}

\usepackage{amsmath}
\usepackage{amssymb}
\usepackage{mathtools, nccmath}
\usepackage{amsfonts}
\usepackage{nicefrac}
\usepackage[dvipsnames]{xcolor}
\usepackage{colortbl}
\usepackage{algorithm}
\usepackage[noend]{algorithmic}
\usepackage[normalem]{ulem}
\usepackage[format=plain,
            labelfont=it,
            labelsep=period]{caption}
\usepackage{subcaption}
\usepackage{booktabs}
\usepackage{comment}
\usepackage{picins}
\usepackage{floatflt}
\usepackage{pbox}

\definecolor{nhred}{rgb}{0.85,0.25,0.2}
\definecolor{nhgreen}{rgb}{0.1, 0.5, 0.25}

\definecolor{citecolor}{HTML}{0b64c5}
\newcommand{\colorcellh}{\cellcolor{citecolor!15}}

\title{\LARGE \bf
MoDem-V2: Visuo-Motor World Models \\ 
    for Real-World Robot Manipulation
}

\author{Patrick Lancaster$^{1}$ Nicklas Hansen$^{2}$ Aravind Rajeswaran$^{1}$ Vikash Kumar$^{1}$ 
\thanks{$^{1}$Patrick Lancaster, Aravind Rajeswaran, and Vikash Kumar are with Meta AI {\tt\small \{plancaster, aravraj\}@meta.com, vikashplus@gmail.com}}%
\thanks{$^{2}$Nicklas Hansen is with the University of California San Diego, {\tt\small nihansen@ucsd.edu}}%
}

\begin{document}

\maketitle

\thispagestyle{empty}
\pagestyle{empty}

\begin{abstract}
Robotic systems that aspire to operate in \textit{uninstrumented real-world environments} must perceive the world directly via onboard sensing. Vision-based learning systems aim to eliminate the need for environment instrumentation by building an implicit understanding of the world based on raw pixels, but navigating the contact-rich high-dimensional search space from solely sparse visual reward signals significantly exacerbates the challenge of exploration.  The applicability of such systems is thus typically restricted to simulated or heavily engineered environments since agent exploration in the real-world without the guidance of explicit state estimation and dense rewards can lead to unsafe behavior and safety faults that are catastrophic. In this study, we isolate the root causes behind these limitations to develop a system, called MoDem-V2, capable of learning contact-rich manipulation directly in the uninstrumented real world. Building on the latest algorithmic advancements in model-based reinforcement learning (MBRL), demo-bootstrapping, and effective exploration, MoDem-V2 can acquire contact-rich dexterous manipulation skills directly in the real world. We identify key ingredients for leveraging demonstrations in model learning while respecting real-world safety considerations -- exploration centering, agency handover, and actor-critic ensembles. We empirically demonstrate the contribution of these ingredients in four complex visuo-motor manipulation problems in both simulation and the real world. To the best of our knowledge, our work presents the first successful system for demonstration-augmented visual MBRL trained directly in the real world. Visit \webpage{} for videos and more details.

\end{abstract}

\section{Introduction}
\label{sec:intro}

Robot agents learning manipulation skills directly from raw visual feedback avoid the need for explicit state estimation and extensive environment instrumentation for rewards, but face heightened exploration, and thereby safety, challenges in navigating the contact-rich high-dimensional search space purely based on sparse visual reward signals. These challenges are especially critical for agents operating in the real world where inefficiency can be expensive, and safety faults can be catastrophic. One approach to developing robot manipulation policies that avoid such safety restrictions is simulation to reality transfer~\cite{Rusu2016SimtoRealRL, Tobin2017DomainRF, Kumar2021RMARM, Handa2022DeXtremeTO}. However, the creation and calibration of accurate physics simulations (from first principles) for contact-rich tasks is extremely challenging and time-consuming. In this work, we alternatively study the use of visual world model learning~\cite{ha2018worldmodels, hafner2019planet, hansen2022temporal} for robot manipulation directly from real-world interaction.

\def\firstwidth{0.23}
\begin{figure}[t]
     \centering
     \begin{subfigure}[b]{\firstwidth\textwidth}
         \centering
         \includegraphics[width=\linewidth]{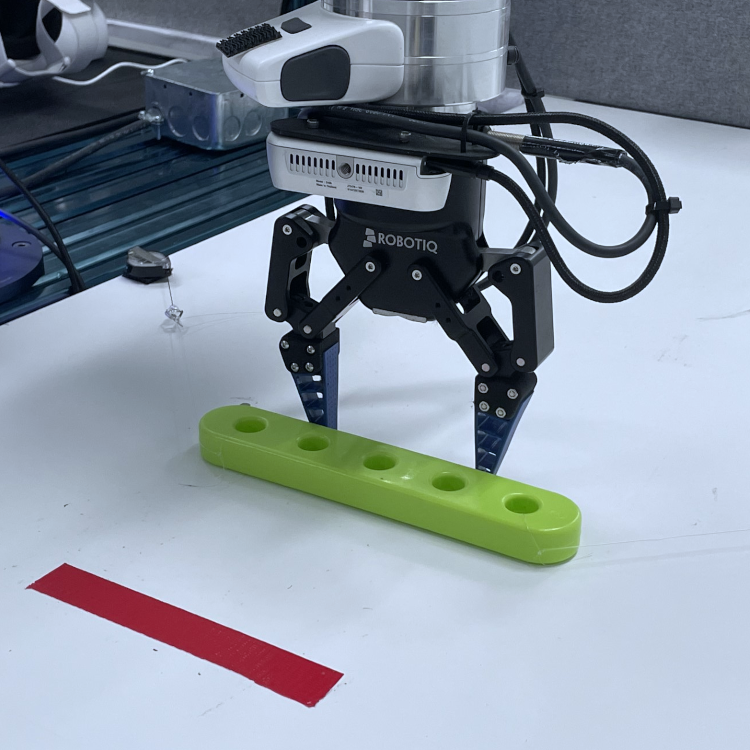}
     \end{subfigure}
     \begin{subfigure}[b]{\firstwidth\textwidth}
         \centering
         \includegraphics[width=\textwidth]{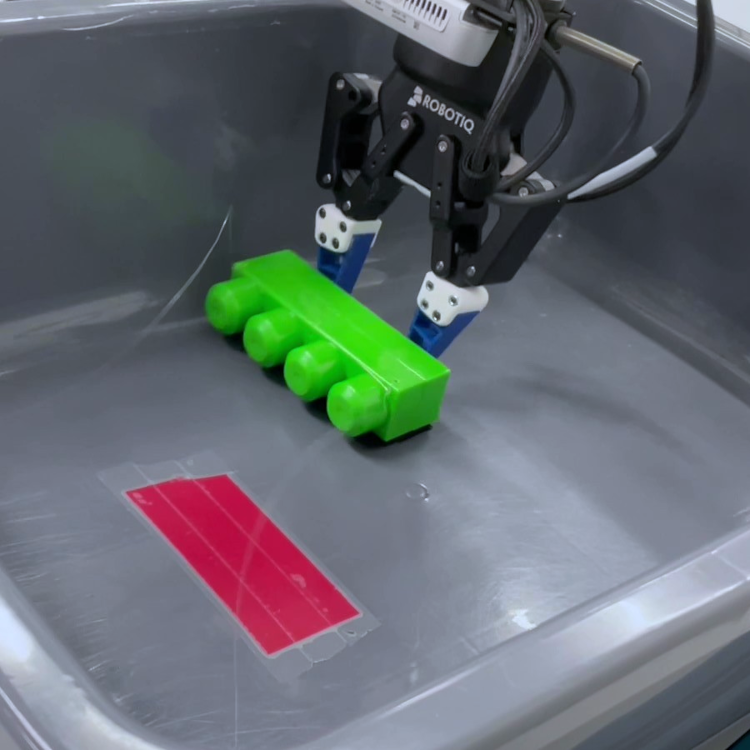}
     \end{subfigure}
     \par\smallskip
     \begin{subfigure}[b]{\firstwidth\textwidth}
         \centering
         \includegraphics[width=\textwidth]{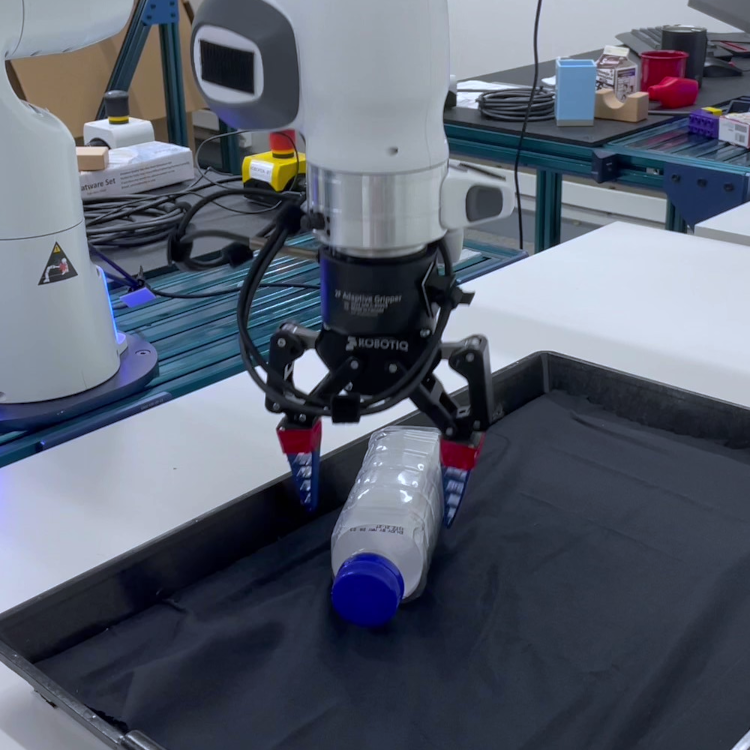}
     \end{subfigure}
     \begin{subfigure}[b]{\firstwidth\textwidth}
         \centering
         \includegraphics[width=\textwidth]{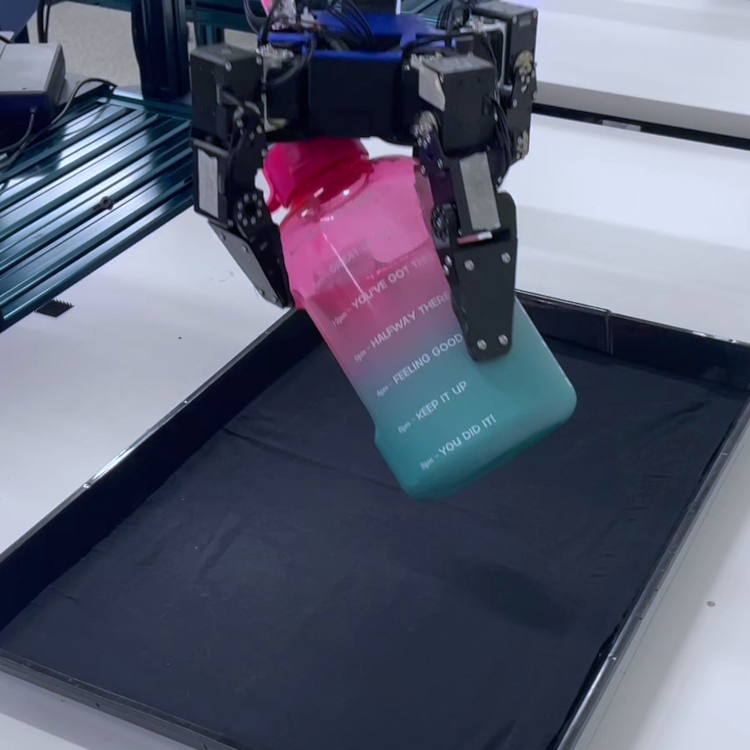}
     \end{subfigure}     
        \caption{\footnotesize{We use MoDem-V2 to train the robot on four contact-rich manipulation tasks. These tasks cover a wide range of manipulation skills, namely non-prehensile pushing, object picking, and in-hand manipulation. In recognition of the difficulty of robust pose tracking and dense reward specification in the real world, the robot performs these tasks using only raw visual feedback, proprioceptive signals, and sparse rewards.}}
        \label{fig:tasks}
\end{figure}

Model-Based Reinforcement Learning~(MBRL) with visual world models involves the learning of dynamics models using real-world data, directly from visual observations. When applied to robot manipulation, visual MBRL can mitigate the need for detailed physics simulations from first principles, as well as the need for specialized sensor instrumentation and state estimation pipelines. However, visual MBRL for real-world robotics still has two major challenges: (a) sample inefficiency; and (b) sparse/weakly shaped rewards. While a number of recent algorithms such as RRL \cite{shah2021rrl} and MoDem \cite{hansen2022modem} circumvent these challenges by leveraging a small number of expert demonstrations to improve sample-efficiency, they rely on aggressive exploration to compensate for weak reward supervision that can result in unsafe behaviors, restricting their application to simulated or heavily engineered scenarios. 

\begin{figure}[t]
     \includegraphics[width=\linewidth]{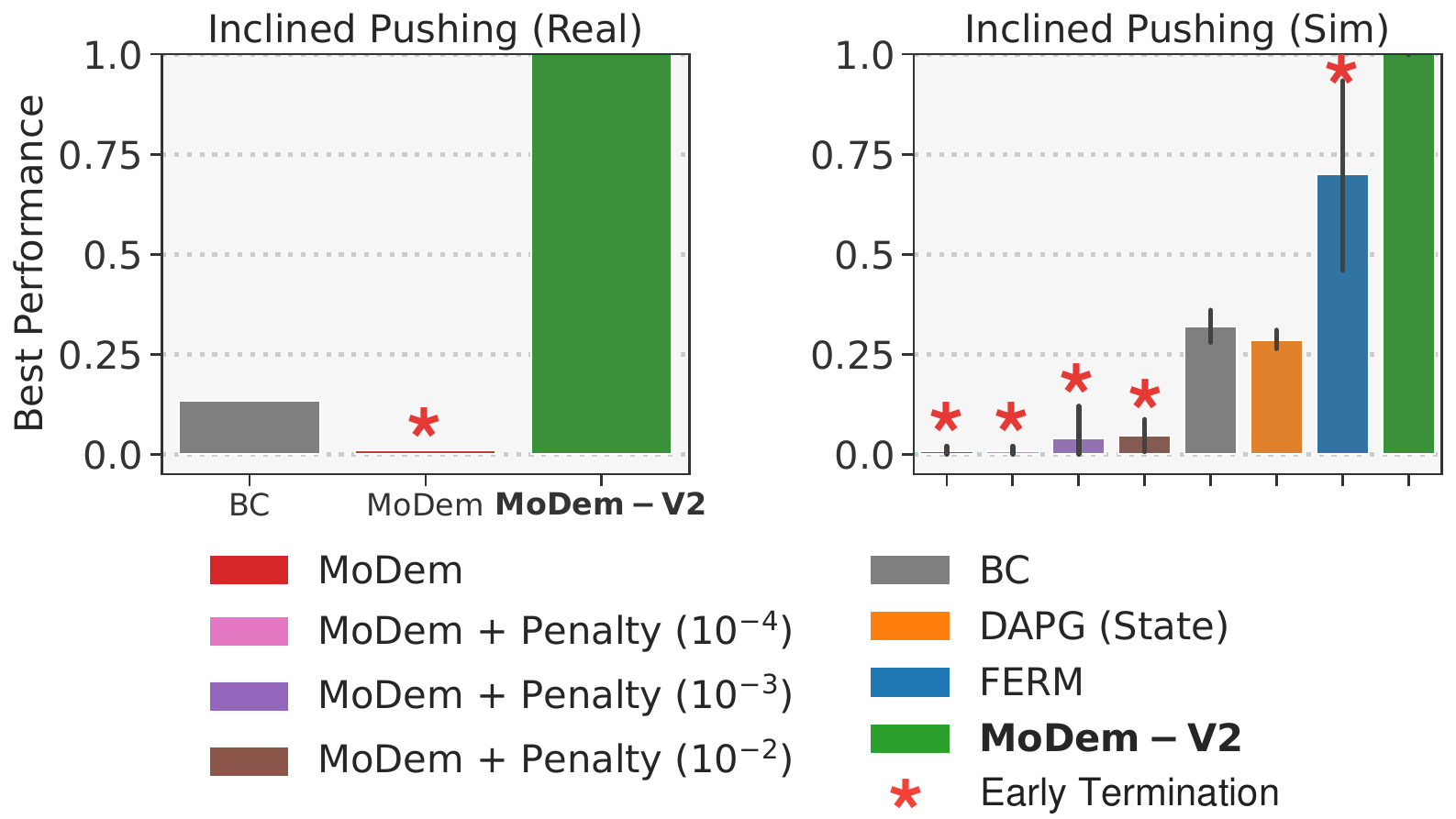}
    \caption{Agent performance on the inclined pushing task before failure due to safety violations. An asterisk indicates that agent training was terminated due to significant safety violations. \emph{Left:} On a real-world robot, MoDem violates (robot manufacturer specified) torque limits at the onset of online interaction and is unable to learn, whereas MoDem-V2's conservative exploration allows it to perfect the task. \emph{Right:} Further evaluation in simulation reveals that simply penalizing the amount of torque exerted by the robot does not prevent termination due to significant safety violations. Other baseline agents are either terminated due to unsafe behavior or achieve significantly lower success than MoDem-V2.}
    \label{fig:torque_penalty}
\end{figure}

We indeed found MoDem to be infeasible for direct application in the real world due to excessively aggressive exploratory behavior. Even with significant engineering investments for safety, we found that the built-in low-level hardware controllers/drivers fault repetitively owing to excessive velocity, acceleration, and torque in the robot’s joints that exert dangerous amounts of force on the environment or the robot itself. While some level of safety can be imposed through hard-coded limits on velocity, acceleration, and torque (as done in all of our experiments), this is an insufficient solution for preventing unsafe behavior during online interaction (\autoref{fig:torque_penalty} left). Tuning these task-specific limits for interaction in the real world is costly and faced with operational and safety challenges. Balancing the tightness of these limits presents a further challenge, as aggressive limits can prohibit the robot from exerting the minimum energy needed to solve the task and weak limits will fail to inhibit unsafe actions. Finally, these static limits fail to incorporate the need for dynamic risk sensitivities across relevant time scales (such as at the scale of a single episode, or even at the scale of training epochs). Furthermore, simply penalizing the amount of torque exerted by the agent is an ineffective, retrospective solution that does not prevent unsafe actions at the onset of exploration as shown in \autoref{fig:torque_penalty} and \autoref{fig:torque_penalty_series}. How can we get around these challenges?

Our key insight is that conservative exploration can respect the safety constraints of real-world environments while still allowing the agent to modulate its strategy according to task progress such that it is able to learn quickly and efficiently. We translate this insight into implementation in three steps. First, rather than sampling actions from the entire action space, \textbf{warm-starting} exploration with actions sampled from a policy learned via behavioral cloning (BC) ~\cite{Atkeson1997RobotLF} prevents the agent from straying far from the provided demonstrations at the onset of online learning. Second, as our world model gains better coverage through online exploration, \textbf{agency-transfer} gradually shifts the agent from executing BC policy actions to short-horizon planning. Agency transfer provides a mechanism for increased exploration while stymying over-optimistic evaluation of regions of the observation-action space far from the agent’s previous experience. Third, we use \textbf{actor-critic ensembles} to estimate the epistemic uncertainty of the value estimations of these short-horizon trajectories, allowing the agent to avoid overly optimistic actions. We integrate these three enhancements into the recently proposed MoDem \cite{hansen2022modem} algorithm to develop MoDem-V2. Despite the simplicity of the individual components, their combination, and the resulting effectiveness in transforming overly aggressive, fault-prone MoDem agents into MoDem-V2 agents that efficiently and safely learn manipulation behaviors in the real-world is quite unique. Our work is, to the best of our knowledge, the first successful demonstration of demonstration-augmented visual MBRL trained directly in the real world.

To evaluate the effectiveness of our approach, we study four robot manipulation tasks from visual feedback (\autoref{fig:tasks}), and their simulated counterparts. 

 Our main contributions are:

\begin{itemize}
\itemsep0em
\item We \textbf{identify unsafe exploration and over-optimism} as the key issues in leveraging visual MBRL algorithms for real-world applications.
\item  From this insight, we develop MoDem-V2 by integrating three ingredients into MoDem, namely \textbf{policy centering}, \textbf{agency transfer}, and
\textbf{actor-critic ensembles}.
\item We demonstrate that MoDem-v2’s conservative exploration significantly enhances its safety profile compared to other baselines, while still retaining the sample-efficient learning capability of MoDem.

\item Finally, we demonstrate that MoDem-V2 can \textbf{quickly learn a variety of contact-rich manipulation skills}, such as pushing, picking, and in-hand manipulation directly in the real world.
\item We contribute towards lowering the barrier of entry for RL in the real world by \textbf{open-sourcing our implementation of MoDem-V2} and discussing practical considerations for training on real hardware.
\end{itemize}

\section{Preliminaries}
\label{sec:preliminaries}
We begin by introducing notations and providing an overview of MBRL settings.

\textbf{Notation:} The general setting of an agent interacting with its environment can be formulated as a Markov Decision Process (MDP) described by the tuple $\mathcal{M} := (\mathcal{S}, \mathcal{A}, \mathcal{T}, R, \gamma)$. Here, $\mathcal{S}$ denotes the state space, $\mathcal{A}$ is the action space, the conditional probability distribution $\mathbf{s}_{t+1} \sim \mathcal{T}(\cdot | \mathbf{s}_{t}, \mathbf{a}_{t})$ defines the dynamics of the MDP, and a scalar reward function is given by $r_{t} = R(\mathbf{s}_{t}, \mathbf{a}_{t})$. Finally,  $\gamma \in [0,1)$ defines the discount factor for the MDP to trade-off future rewards to current ones. The goal for an agent is to learn a policy $\pi: \mathcal{S} \mapsto \mathcal{A}$ that can achieve high long term performance given by $\mathbb{E}_\pi \left[ \sum_{t=0}^\infty \gamma^t r_t \right]$.

We specifically consider the problem of robotic manipulation from visual feedback on real hardware. We aim to learn a control policy that controls a physical robot from RGB observations provided by cameras placed in the scene, and robot proprioception. We model this setting as a \textbf{high-dimensional} MDP with \textbf{sparse rewards.} This assumes that while the state space of the MDP (\emph{e.g.} object poses) is not directly observable by the agent, a sufficient representation of the state can be well-approximated through the combination: $\mathbf{s} = (\mathbf{x}, \mathbf{q})$ where $\mathbf{x}$ denotes stacked RGB observations from the robot's camera(s) and $\mathbf{q}$ denotes proprioception from the robot. Finally, we only assume access to a sparse task completion reward, which is much easier to obtain via visual inputs compared to a detailed well shaped reward function. The final goal is to learn a policy that achieves high performance, using minimal online interactions, while respecting the safety considerations of hardware.

\textbf{MoDem - Model-Based Reinforcement Learning with Demonstrations:}
Our approach is based on \emph{MoDem} \cite{hansen2022modem} -- a MBRL algorithm that combines \emph{(i)} model predictive control (MPC) and the decoder-free world model of TD-MPC \cite{hansen2022temporal} with \emph{(ii)} a small number of demonstrations to efficiently solve continuous control problems with limited online interaction. Concretely, MoDem learns the following five components:
\begin{equation}
    \label{eq:tdmpc-components}
    \begin{array}{lll}
        \text{State embedding} & \mathbf{z} = h_{\theta}(\mathbf{s})\\
        \text{Latent dynamics} & \mathbf{z}' = d_{\theta}(\mathbf{z}, \mathbf{a}) \\
        \text{Reward predictor} & \hat{r} = R_{\theta}(\mathbf{z}, \mathbf{a}) \\
        \text{Terminal value} & \hat{q} = Q_{\theta}(\mathbf{z}, \mathbf{a}) \\
        \text{Policy guide} & \hat{\mathbf{a}} = \pi_{\theta}(\mathbf{z}) 
    \end{array}
\end{equation}
where $h_{\theta},d_{\theta},R_{\theta}$ and $Q_{\theta}$ are learned end-to-end using a combination of joint-embedding predictive learning \cite{Grill2020BootstrapYO}, reward prediction, and Temporal Difference (TD) learning, and $\pi_{\theta}$ is a deterministic policy that learns to maximize $Q_{\theta}$ conditioned on a latent state $\mathbf{z}$ (see \autoref{eq:tdmpc-loss} of the Appendix). Throughout this work, we will refer to $(h_{\theta},d_{\theta},R_{\theta},Q_{\theta})$ as the \emph{world model}, and $\pi_{\theta}$ as the \emph{policy}.  See \cite{hansen2022temporal,hansen2022modem} for additional details on the world model. While \emph{MoDem} has been shown to be effective in simulation, owing to aggressive exploration its applicability in domains where safety constraints can't be overlooked is limited (\autoref{fig:torque_penalty}).

 \section{Safety}
 \label{sec:safety}
Agents that seek to learn via continuous operations must respect hardware and environmental safety constraints. Such constraints are diverse, obscure, and unobservable (intrinsic to low-level hardware details, or lack of appropriate sensing) and therefore cannot be directly accounted for by the agent during operations. Alternatives such as action penalization and user-defined safety are also insufficient (\autoref{fig:torque_penalty}) requiring extensive human intervention and monitoring for operations. In this work, we refer to these unobservable constraints as \textbf{safety violations}. For real-world operations, they are defined as hardware faults that require human intervention. In simulation, we define them as violations (unobserved by the policy) in either the robot’s torque limits (as defined by the robot manufacturer) or excessive contact force (100 N) applied by the robot’s end effector.

\section{Method}
\label{sec:method}

\begin{figure}[t!]
\vspace{-0.15in}
\begin{algorithm}[H]
\caption{~~Planning procedure of MoDem-V2 \\ 
(\textcolor{nhred}{$\bullet$Original MoDem}~~\textcolor{nhgreen}{$\bullet$MoDem-V2 modification})}
\label{alg:inference}
\begin{algorithmic}[1]
\REQUIRE $\theta:$ learned network parameters \\ 
~~~~~~~$\mu, \sigma$: initial parameters for $\mathcal{N}$\\
~~~~~~~$N$: num sample  trajectories \\ 
~~~~~~~$\mathbf{s}_{0}, h$: current state, planning horizon \\
~~~~~~~$\tau$: trajectory weighting temperature \\
~~~~~~~{\color{nhgreen}$\alpha$: probability of using model rollouts} \\
~~~~~~~{\color{nhgreen}$w_1, w_2$: ensemble mixing weights}
\STATE encode state $\mathbf{z}_{0} \leftarrow h_{\theta}(\mathbf{s}_{0})$~~~~~~~~~~~$\vartriangleleft$ {\color{CadetBlue}\emph{State embedding}}
\IF {{\color{nhgreen} $\operatorname{rand}() > \alpha$}}
\STATE {\color{nhgreen} $\Gamma \coloneqq \{\mathbf{a}_{0}\}^N \sim$ $\pi^{BC}_{\theta}$}~~~~$\vartriangleleft$ {\color{CadetBlue}\emph{Center actions around BC}}
\STATE {\color{nhgreen} $\phi_{\Gamma} =  Q^{BC}_{\theta}(\mathbf{z}_{0}, \{\mathbf{a}_{0}\}^N)$}~~~~~~~~~~~{\color{CadetBlue}$\vartriangleleft$ \emph{Critic evaluation}}
\ELSE
\STATE { \color{nhred} \sout{$\Gamma \coloneqq\mathbf{a}_{t:t+h} \sim  \mathcal{N}(\mu, \mathrm{I}\sigma^{2})$}}~~~~~~~~~~~~$\vartriangleleft$ {\color{CadetBlue}\emph{Prior sampling}}
\STATE {\color{nhgreen}  $\Gamma \coloneqq \mathbf{a}_{t:t+h} \sim \pi^{1:M}_{\theta}, d_{\theta}$}$\vartriangleleft$ {\color{CadetBlue}\emph{Policy ensemble sampling}}
\FOR{all $N$ trajectories $\Gamma_i =(\mathbf{a}_{t}, \mathbf{a}_{t+1},\dots, \mathbf{a}_{t+h})$}
\FOR{step $t=0..h-1$}
\STATE $\phi_{\Gamma} = \phi_{\Gamma} + \gamma^{t} R_{\theta}(\mathbf{z}_{t}, \mathbf{a}_{t})$~~~~~~~~~~~~~~~~{\color{CadetBlue}$\vartriangleleft$ \emph{Reward}}
\STATE $\mathbf{z}_{t+1} \leftarrow d_{\theta}(\mathbf{z}_{t}, \mathbf{a}_{t})$~~~~~~~~~~~~{\color{CadetBlue}$\vartriangleleft$ \emph{Latent transition}}
\ENDFOR
\STATE {\color{nhred} \sout{$\phi_{\Gamma} = \phi_{\Gamma} + \gamma^{h} Q_{\theta}(\mathbf{z}_{H}, \mathbf{a}_{H})$}}~~~~~~~{\color{CadetBlue}$\vartriangleleft$ \emph{Terminal value}}
\STATE {\color{nhgreen} \# Ensemble of terminal values}
\STATE {\color{nhgreen} $\phi^{1:M}_{\Gamma} = \phi_{\Gamma} + \gamma^{h} Q^{1:M}_{\theta}(\mathbf{z}_{H}, \mathbf{a}_{H})$}
\STATE {\color{nhgreen} \# Epistemic uncertainty estimation}
\STATE {\color{nhgreen}  $\phi_{\Gamma} = w_1$ mean$(\phi^{1:M}_{\Gamma}) + w_2$ std $(\phi^{1:M}_{\Gamma})$}
\ENDFOR
\ENDIF
\STATE $\Omega = e^{\tau (\phi_{\Gamma})}$, $\mu = \frac{\sum_{i=1}^{N} \Omega_{i} \Gamma_{i}}{\sum_{i=1}^{N} \Omega_{i}}$, $\sigma = \sqrt{\frac{\sum_{i=1}^{N} \Omega_{i} (\Gamma_{i} - \mu)^{2}}{\sum_{i=1}^{N} \Omega_{i}}}$
\STATE \textbf{return} $\mathbf{a} \sim \mathcal{N}(\mu, \mathrm{I}\sigma^{2})$
\end{algorithmic}
\end{algorithm}
\vspace{-0.3in}
\end{figure}

In this work, we aim to learn manipulation skills through environment interaction on \emph{real} robots, from \emph{visual} feedback, and with \emph{minimal} human supervision and intervention. We first discuss 
two important metrics for any agent that aims to quickly and safely learn manipulation skills in the real world. We then address these limitations with three proposed enhancements to MoDem in order to develop MoDem-V2.

\textbf{Strengths and Weaknesses of MoDem:}  MoDem accelerates learning through a three-stage framework in which $h_{\theta},\pi_{\theta}$ are first pretrained on a set of demonstrations using Behavior Cloning (BC), and the resulting policy $\pi_{\theta} \circ h_{\theta}(\cdot)$ is then used to \emph{seed} the model, \emph{i.e.} collect a small initial dataset for learning the model (with Gaussian noise injected into $\pi_{\theta}$ for exploration). After initializing the model on seeding data, the world model is iteratively used to collect new data via online interaction, and is optimized on all data: demonstrations, seeding data, and online interaction data.  Using these insights, MoDem demonstrates accelerated model learning in a collection of simulated tasks. Yet, when exposed to the real world,  MoDem faces a variety of challenges that were suppressed during its original simulation study. 

We find that MoDem exerts excessive forces and torques far beyond those exemplified in the provided demonstrations (see  \autoref{sec:experiments}).  Even at the beginning of the online interaction phase at which the agent has only observed data close to the BC policy, MoDem relies on its world model and value function to discriminate between high and low reward actions. Thus when MoDem samples actions across the entire action space, the world model and value function cannot (at least initially) provide good estimates. This can result in task failure due to poor action selection, or lead to the robot exerting unsafe forces and torques by choosing consecutive actions that are far apart in action space. Our primary contribution lies in eliminating these limitations while also improving the effectiveness of the original method.

\subsection{MoDem-V2: Real-World Model-Based Reinforcement Learning with Demonstrations}
\label{sec:approach}
MoDem generates agent actions by first mapping raw observations into a learned lower-dimensional space, and then performing efficient short horizon planning in that latent space with its learned dynamics model and value function. We propose the following three adaptations to MoDem's planning procedure in order to improve its safety while maintaining its strengths in autonomy and data-efficiency (Algorithm \ref{alg:inference}).

\textbf{Policy centered actions:} Rather than sampling actions from across the entire action space, we propose to sample actions from our learned policy. This more conservative exploration strategy reduces the likelihood of world model and value function evaluation over unseen regions of the state-action space, enabling them to better discriminate the quality of the generated actions.

\textbf{Agency transfer from BC actions to MPC:} At the beginning of the online interaction phase, MoDem immediately begins using its learned world model and value function to do MPC. Yet both of these components have only seen limited data near the BC policy, so relying on them to choose actions for multiple consecutive timesteps at the beginning of interaction can quickly lead the agent into an unexplored region of the observation-action space from which it cannot recover. Our remedy is to \textit{gradually} shift from executing actions sampled from the BC policy to actions computed by short horizon planning. We implement this in Algorithm \ref{alg:inference} with a hyperparameter $\alpha$ that is initialized to 0 at the beginning of interaction and linearly increases to 1.0 over a fixed number of interaction steps.

\textbf{Actor-Critic Ensembles for uncertainty aware planning:} The use of actor-critic ensembles \cite{huang2017learning} improves the agent’s value estimations in two primary ways. First, note that each actor is trained by optimizing it to maximize its corresponding critic. While this provides a solution for efficiently finding the maximum value of Q over actions, it is subject to significant overestimation bias \cite{fujimoto2018addressing}. We mitigate this by only evaluating a critic with final trajectory actions produced by policies not  directly optimized to maximize that particular critic. Actor-critic ensembles also improve value estimation by providing the agent with a pool of independently trained value functions, each of which computes its own value estimate. By estimating the epistemic uncertainty \cite{chua2018deep} of a trajectory, the agent can make uncertainty-aware decisions. We incorporate this into MoDem-V2 with weights $w_1 > 0$ and $w_2 < 0$ in Algorithm \ref{alg:inference}.

\section{Experimental Design}

We design experiments to evaluate the design choices behind MoDem-V2 in enabling real-world contact-rich manipulation. Our investigation focuses on the following directions:

\begin{itemize}
    \item How sample-efficient is MoDem-V2 relative to other methods?
    \item Is MoDem-V2 safer (\emph{i.e.} fewer safety violations) than other methods including MoDem?
    \item Does MoDem-V2 actually enable physical robots to learn real-world manipulation tasks?
\end{itemize}

Our first step in answering these questions is to use simulation \cite{robohive} to compare our method to both the original MoDem and other strong reinforcement learning baselines that also uses demonstrations to guide policy learning, Demonstration Augmented Policy Gradients (DAPG) \cite{Rajeswaran-RSS-18} and the Framework for Efficient Robot Manipulation (FERM) \cite{Zhan2020AFF}. We also provide further analysis of MoDem-V2 by ablating each of its three design decisions. Finally, we deploy MoDem-V2 onto a physical robot and evaluate its capability to learn four different manipulation tasks in the real world.

\label{sec:experimentalDesign}

\subsection{Hardware}

\begin{wrapfigure}{r}{0.32\linewidth}
    \centering
    \vspace{-2em}
    \includegraphics[width=\linewidth]{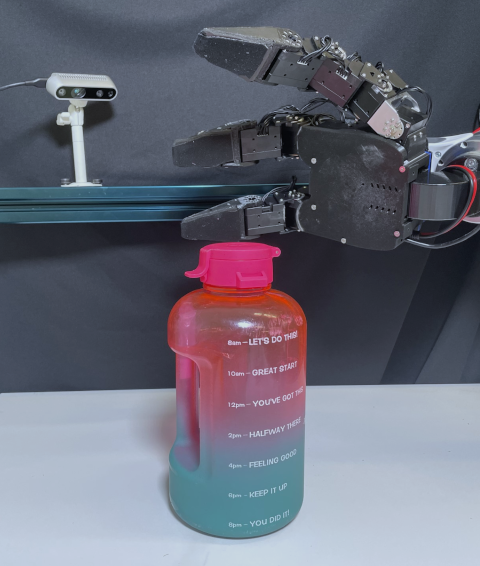}
  \caption{A view of the in-hand reorientation task as an example of our hardware setup.}
  \label{fig:hardware_setup}
\end{wrapfigure}

We adopt a set of core hardware components that are common across all of our experimental tasks. Each task uses a Franka Panda arm. The pushing and picking tasks use a Robotiq two-fingered gripper, while the in-hand reorientation task uses a ten-degree of freedom D'Manus hand \cite{bhirangi2022all} from the \textit{ROBEL} ecosystem \cite{ahn2020robel}{}. For perception, three RealSense D435 cameras are mounted to the left, right and above the robot. Our hardware setup is depicted in \autoref{fig:hardware_setup}.

\subsection{Task Suite}
We evaluate MoDem-V2 on four manipulation tasks in both simulation and the real world. These tasks encompass a variety of manipulation skills, namely pushing, picking, and in-hand manipulation as shown in \autoref{fig:tasks}. We briefly describe each task below, see \autoref{sec:add_task_details} for more details.  

\textbf{Planar Pushing:} This task requires the robot to push an oblong object towards a fixed goal position on a table top.  This task is likely the easiest of all four tasks, and we view it as base case with which to compare the other tasks.

\textbf{Inclined Pushing:} This task requires the robot to push an object up an incline to reach a fixed goal position. During execution of the task, the robot must raise its gripper such that it can progress up the incline while also making sufficiently precise contact with the block to prevent it from slipping beneath or around the side of the gripper.

\textbf{Bin Picking:} To complete this task, the robot must grasp a juice container and then raise it out of the bin.  This task requires accurate positioning of the gripper because the (mostly) non-deformable container has a primary width that is approximately 65\% of the gripper's maximum aperture. This task also requires the robot to disambiguate spatially similar states; \emph{e.g.} if the gripper is above the bin, the robot must understand whether or not the object is in its grasp so that it can decide to go down towards the bin to pick up the object or stay where it is in order to receive reward.

\textbf{In-Hand Reorientation:} This task requires the robot to grasp a water bottle laying on its side and then in-hand manipulate it to an upright position. Using the multi-fingered D’Manus hand more than doubles the dimensionality of the action space relative to the previous tasks. 

\section{Experiments}
\label{sec:experiments}

In this section, we evaluate MoDem-V2 against strong baselines in simulation and measure its performance in real world environments. Our simulation experiments measure both sample-efficiency and safety (defined in \autoref{sec:safety}), two important aspects for any learning method that is used in the real world. We then use MoDem-V2 to train a robot to perform all four manipulation tasks in the real world. All experiments train an initial policy with only ten demos, and each evaluation is aggregated over 30 trials.

\begin{figure}[t]
  \includegraphics[width=\linewidth]{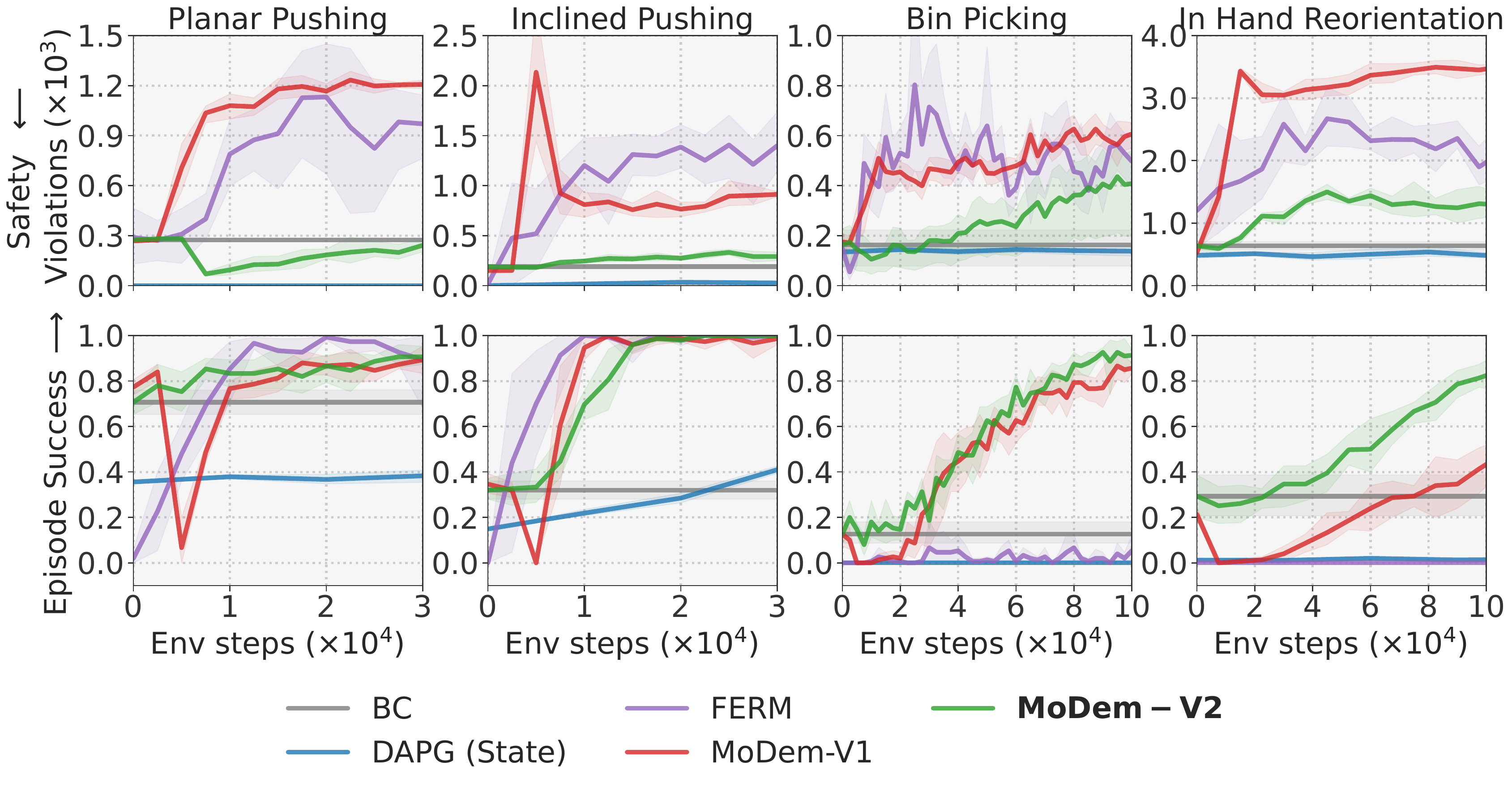}
  \caption{The number of safety violations as defined in \autoref{sec:safety} (top row) and success rate (bottom row) for each of the four manipulation tasks in simulation. Lower is better for safety violations while higher is better for episode success. While both MoDem-V2 and MoDem achieve similar or better sample-efficiency than all of the baselines, MoDem-V2 exhibits significantly safer learning as evidenced by the drastically lower amount of safety violations.}
  \label{fig:sim_compare}
\end{figure}

\subsection{Simulated Comparison to Baselines}

With respect to sample-efficiency, MoDem-V2 and MoDem both significantly outperform DAPG (State), as shown in \autoref{fig:sim_compare} (bottom). This is despite DAPG having access to privileged state information (such as the object's pose) and dense rewards (such as rewards based on the distance between the gripper and the object)! Although FERM achieves similar performance to both versions of MoDem on the easier pushing tasks, it is unable to learn the bin picking and in-hand re-orientation tasks. While MoDem achieves similar efficiency to MoDem-V2 on the first three tasks, it suffers a significant performance drop in the early stages of training across all tasks. This coincides with the point at which MoDem begins to aggressively explore the action space, resulting in the robot applying excessive forces/torques beyond that observed in the provided demonstrations.

We measured the number of safety violations that occurred throughout training for MoDem and MoDem-V2, as shown in  \autoref{fig:sim_compare} (top). While both methods initially commit few violations since their BC policies were trained from the same demonstrations, MoDem's safety violations sharply increase as the interaction phase begins. Thanks to our design decisions, the number of violations exerted by MoDem-V2 is generally lower throughout online learning comparatively. Here MoDem-V2 demonstrates that it can achieve similar or better sample-efficiency than MoDem, while committing significantly fewer safety violations and thereby safer behavior.

\begin{figure}[t]
    \centering
  \includegraphics[width=\linewidth]{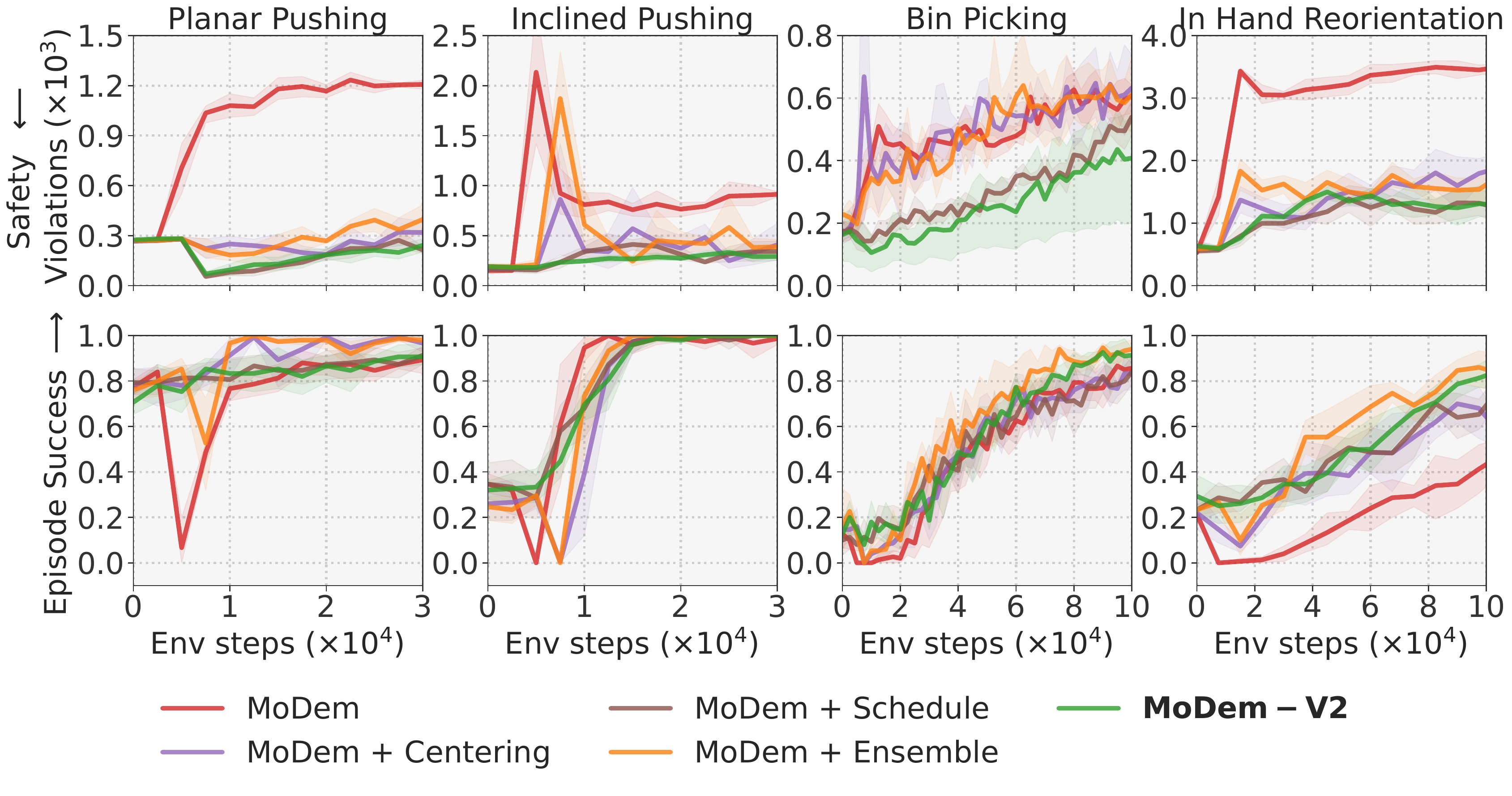}
  \caption{Ablations of the three MoDem-V2 enhancements for all four tasks. Lower is better for safety violations (top row) while higher is better for episode success (bottom row). MoDem-V2 achieves both the higher sample-efficiency of \textit{Ensemble} and the increased safety profile of \textit{Schedule}.}
  \label{fig:ablations}
\end{figure}

\subsection{Ablation of design choices}

We perform ablations of MoDem-V2 by individually adding each improvement specified in \autoref{sec:approach} to MoDem. As shown in \autoref{fig:ablations}, we found that all of the ablations generally maintained or improved over the sample-efficiency of MoDem while significantly improving safety by committing fewer violations. The one exception to this is the bin picking task, for which the \textit{Centering} and \textit{Ensemble} ablations commit a greater number of safety violations whereas MoDem-v2 exhibits much superior performance. When comparing between ablations it is clear that each individual modification has both benefits and drawbacks. First, note that \textit{Centering} is a necessary sub-component of \textit{Schedule}. While \textit{Schedule} is generally safer than \textit{Ensemble}, \textit{Ensemble} typically has better sample-efficiency. By combining these two ingredients, MoDem-V2 is able to achieve the improved sample-efficiency of \textit{Ensemble} while maintaining the improved safety profile of \textit{Schedule}.

\subsection{Real World Results}
As suggested by our simulation experiments, the original MoDem is unsafe for learning real-world manipulation tasks. When we did attempt to run MoDem on our real robot, we found that its aggressive exploration frequently violated safety limits at the beginning of online interaction. For example, on the inclined pushing task, the robot triggered a safety fault within the first two exploration episodes because it exerted excessive force/torque on the incline. Due to the unsafe behavior MoDem induces and the significant human intervention that would be required, it is not feasible to evaluate MoDem in the real world.

In contrast, MoDem-V2 is capable of safely learning manipulation tasks with minimal human intervention. MoDem-V2 enabled the robot to significantly exceed the performance of its initial BC policy with about two hours worth of online training data or less. \autoref{fig:trajectories} shows the success rate of the initial policy cloned from just ten demonstrations and the best MoDem-V2 agent performance achieved throughout online training, as well as example trajectories from the MoDem-V2 agents. Please see \webpage{} for additional videos and results.

\def\realwidth{0.11}
\def\realplotwidth{0.11}
\begin{figure}[t!]
     \centering
     \begin{subfigure}[b]{\realwidth\textwidth}
         \centering
         \includegraphics[width=\linewidth]{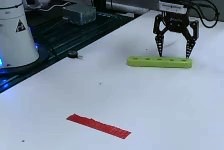}
     \end{subfigure}
     \hfill
     \begin{subfigure}[b]{\realwidth\textwidth}
         \centering
         \includegraphics[width=\textwidth]{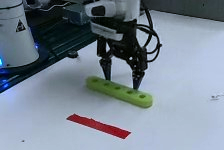}
     \end{subfigure}
     \hfill
     \begin{subfigure}[b]{\realwidth\textwidth}
         \centering
         \includegraphics[width=\textwidth]{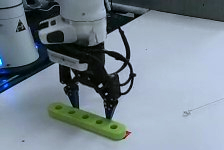}
     \end{subfigure}
     \hfill
     \begin{subfigure}[b]{\realplotwidth\textwidth}
         \centering
         \includegraphics[width=\textwidth]{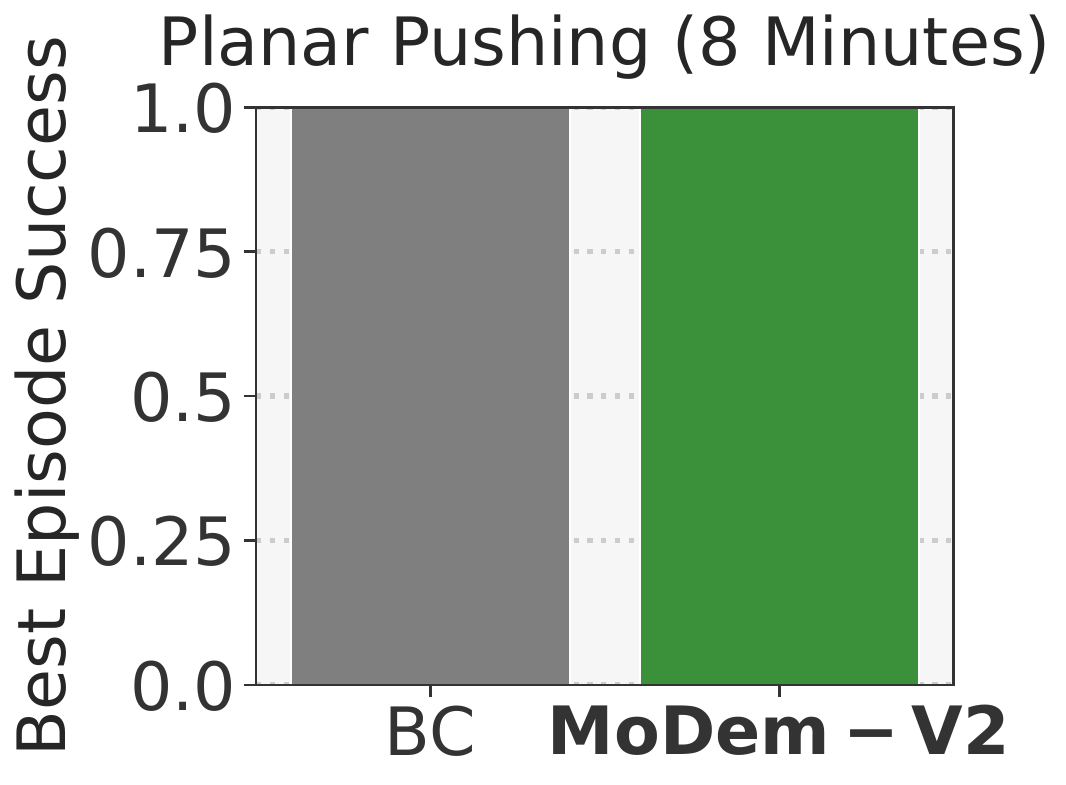}
     \end{subfigure} 
     
     \smallskip
     
     \begin{subfigure}[b]{\realwidth\textwidth}
         \centering
         \includegraphics[width=\linewidth]{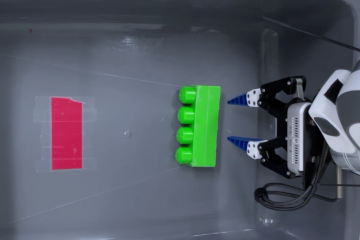}
     \end{subfigure}
     \hfill
     \begin{subfigure}[b]{\realwidth\textwidth}
         \centering
         \includegraphics[width=\textwidth]{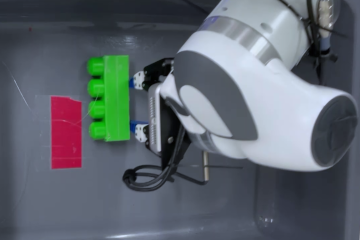}
     \end{subfigure}
     \hfill
     \begin{subfigure}[b]{\realwidth\textwidth}
         \centering
         \includegraphics[width=\textwidth]{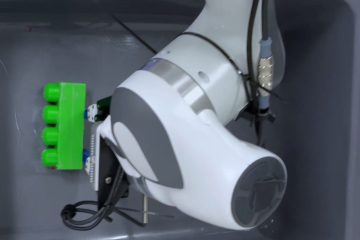}
     \end{subfigure}
     \hfill
     \begin{subfigure}[b]{\realplotwidth\textwidth}
         \centering
         \includegraphics[width=\textwidth]{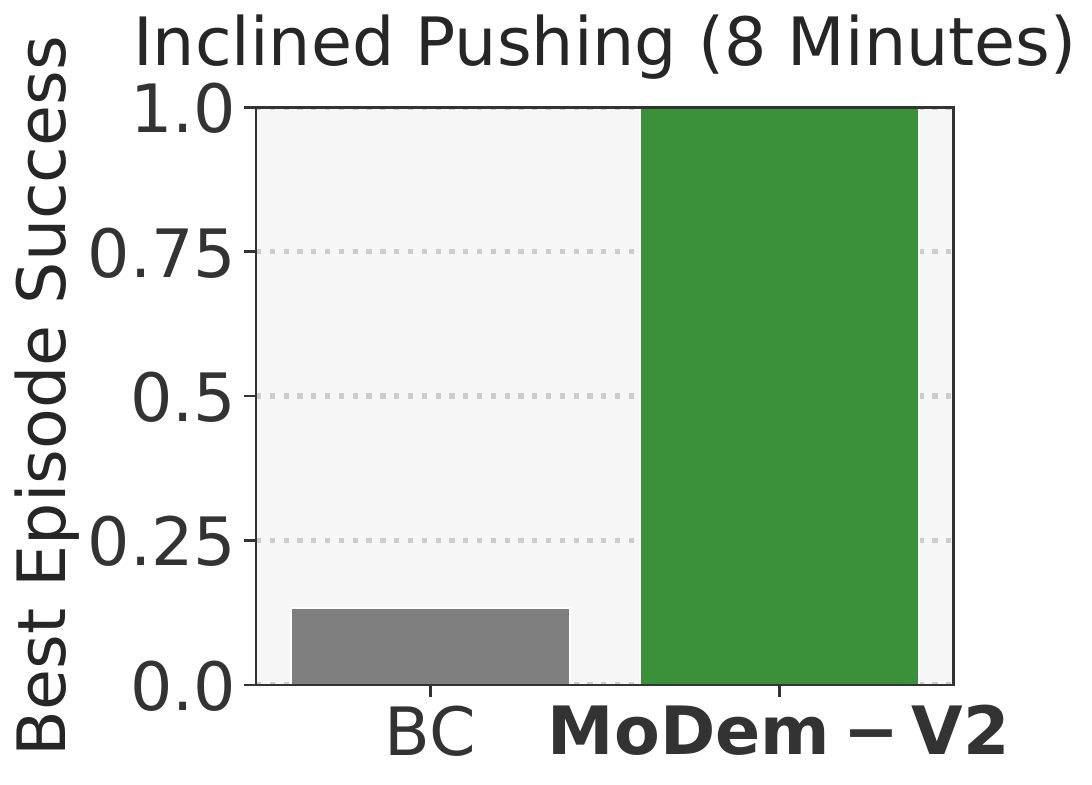}
     \end{subfigure}       

    \smallskip

     \begin{subfigure}[b]{\realwidth\textwidth}
         \centering
         \includegraphics[width=\linewidth]{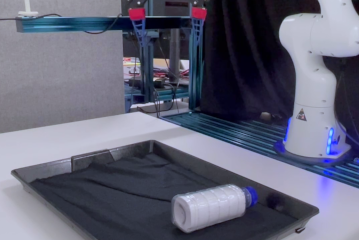}
     \end{subfigure}
     \hfill
     \begin{subfigure}[b]{\realwidth\textwidth}
         \centering
         \includegraphics[width=\textwidth]{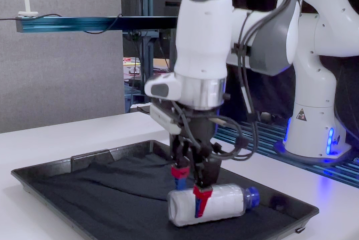}
     \end{subfigure}
     \hfill
     \begin{subfigure}[b]{\realwidth\textwidth}
         \centering
         \includegraphics[width=\textwidth]{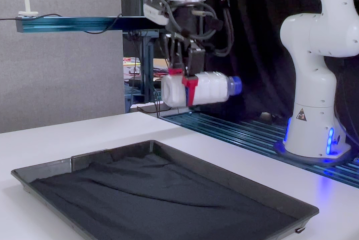}
     \end{subfigure}
     \hfill
     \begin{subfigure}[b]{\realplotwidth\textwidth}
         \centering
         \includegraphics[width=\textwidth]{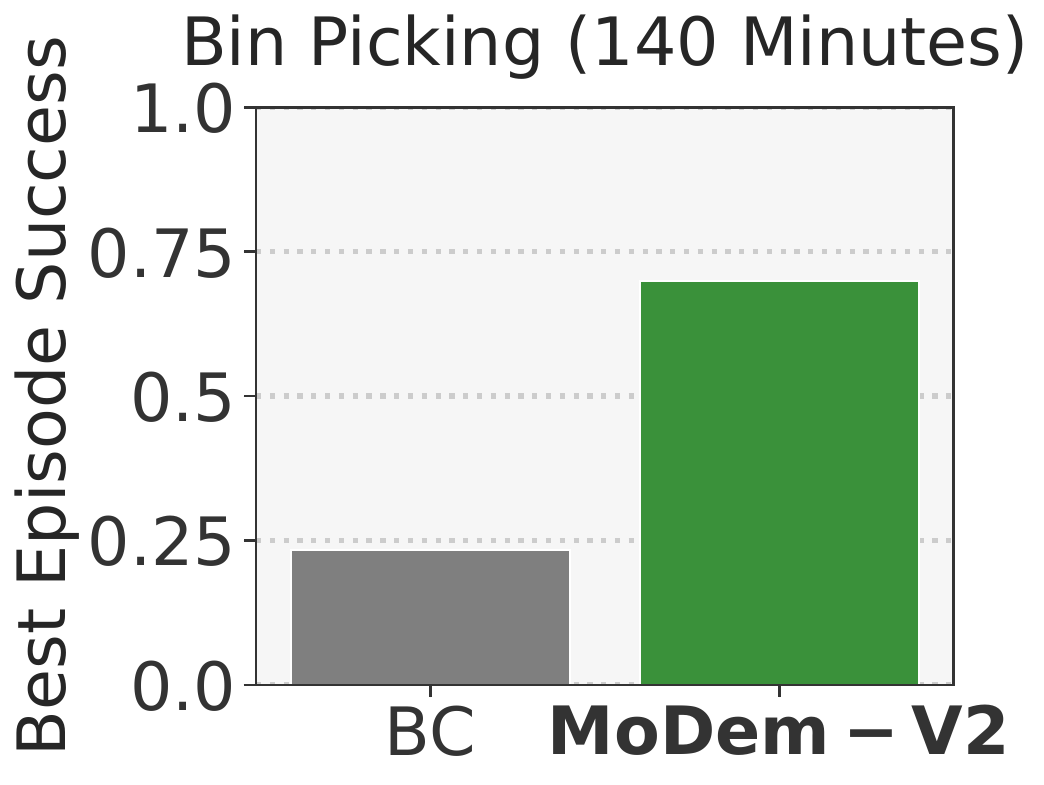}
     \end{subfigure} 
     
     \smallskip
     
     \begin{subfigure}[b]{\realwidth\textwidth}
         \centering
         \includegraphics[width=\linewidth]{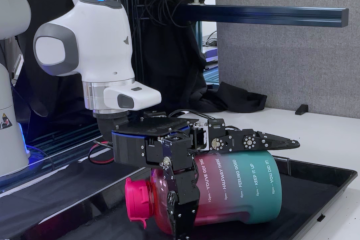}
     \end{subfigure}
     \hfill
     \begin{subfigure}[b]{\realwidth\textwidth}
         \centering
         \includegraphics[width=\textwidth]{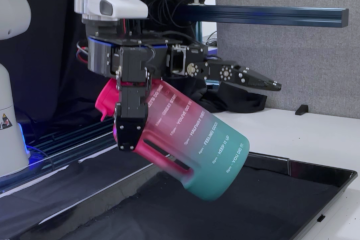}
     \end{subfigure}
     \hfill
     \begin{subfigure}[b]{\realwidth\textwidth}
         \centering
         \includegraphics[width=\textwidth]{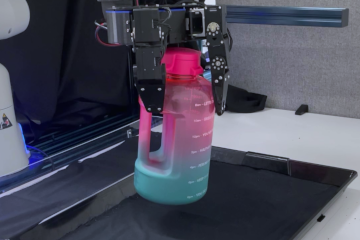}
     \end{subfigure}
     \hfill
     \begin{subfigure}[b]{\realplotwidth\textwidth}
         \centering
         \includegraphics[width=\textwidth]{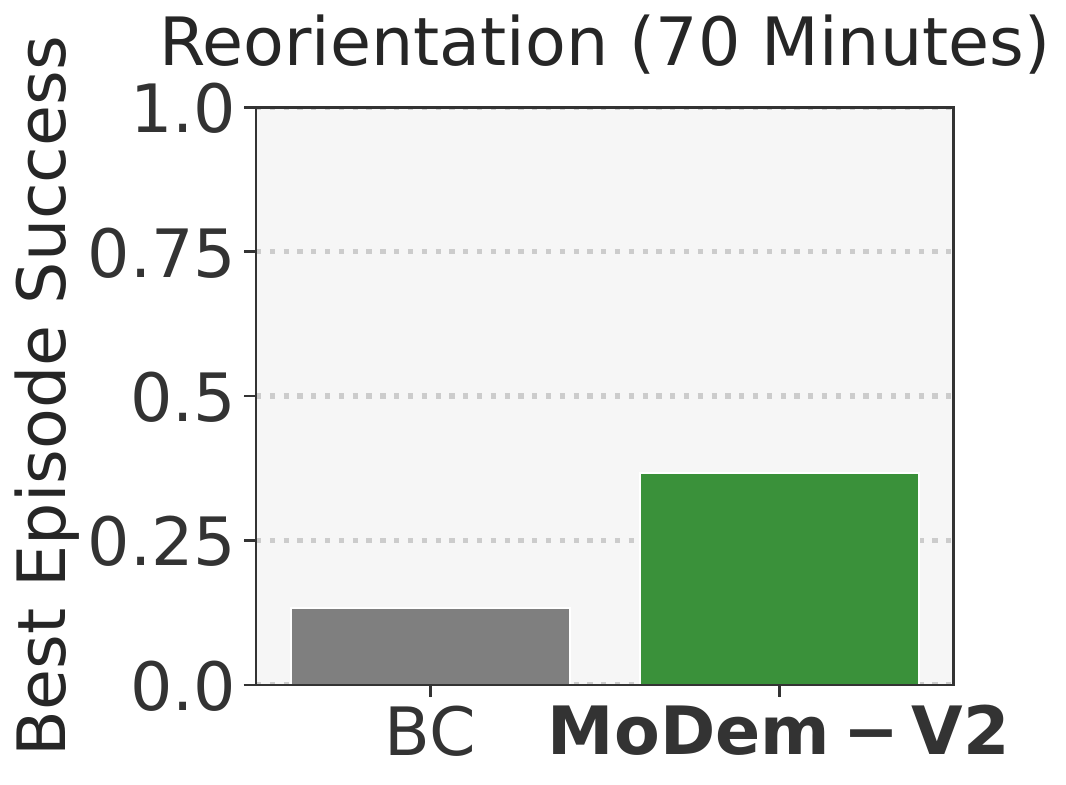}
     \end{subfigure}            
        \caption{\emph{Left:} Example rollouts from our MoDem-V2 agents on real-world manipulation tasks. \emph{Right:} The success rate of our best performing policy and its initial BC policy. MoDem-V2 is able to outperform its BC policy in hours or less. }
        \label{fig:trajectories}
\end{figure}

\section{Related Work}
\label{sec:related-work}

\textbf{Visual MBRL:} Improving sample-efficiency of visual RL by learning a model of the environment has been explored extensively in literature \cite{Ebert2018VisualFM, ha2018worldmodels, hafner2019planet, Kaiser2020ModelBasedRL, Schrittwieser2020MasteringAG, Ye2021MasteringAG, hansen2022temporal}. Here we focus on MBRL algorithms that leverage planning. Prior work typically learns a latent dynamics model from online interaction, and uses a sampling-based planning technique for action selection with candidate action sequences evaluated by the learned model. For continuous control, planning can be formalized as Model Predictive Control (MPC) \cite{Ebert2018VisualFM, hafner2019planet, hansen2022temporal, hansen2022modem}, whereas Monte-Carlo Tree Search (MCTS) is used for discrete action spaces \cite{Schrittwieser2020MasteringAG, Ye2021MasteringAG}. 
Regardless, the majority of work on visual MBRL focus on sample-efficiency in simulated tasks, where practicality and safety are of limited concern. Our work extends the MBRL algorithm of \cite{hansen2022temporal,hansen2022modem} which has already been shown to be very sample-efficient in simulation, and instead focus on the challenges that arise when training MBRL in the real world.

\textbf{Safe Reinforcement Learning:} The field of safe RL encompasses a wide range of  approaches; see \cite{brunke2022safe} for a comprehensive review. A common framework for safe RL is to represent the task as a constrained markov decision process \cite{ge2019safe, chow2018lyapunov, wachi2018safe}, but partial observability presents a challenge to applying such approaches to real world environments. Other methods encode safety as robustness through either domain randomization \cite{tobin2017domain, kontes2020high}  or adversarial perturbation \cite{mehta2020active, yang2022motivating, yang2023stackelberg}. Our work aligns with those previous that used ensemble methods to estimate model uncertainty for guiding the agent towards safer exploration \cite{zhang2020cautious, thananjeyan2020safety, henaff2019model}. Similar to our work, Thananjeyan et. al \cite{thananjeyan2020safety} uses demonstrations to limit policy exploration to be near known safe trajectories. However their method requires the user to provide a function indicating whether a given robot state is safe or not, which can be difficult to specify for high-dimensional or partially observed state spaces. In this work, we focus on proposing solutions for safe exploration that can be deployed on to real robots and do not diminish the high sample efficiency of the original MoDem. 

\textbf{Real-World Robot Learning:} Researchers have explored a wide variety of approaches for robot learning on real hardware, most of which fall into one of three categories: learning from human demonstrations \cite{Jang2022BCZZT, Nair2022R3MAU, Zhu2022VIOLAIL, Brohan2022RT1RT}, learning from large uncurated datasets \cite{Pinto2015SupersizingSL, Levine2016LearningHC, Ebert2018VisualFM}, learning from online interaction via RL \cite{Zhu2018DexterousMW, wu2023daydreamer}, or any combination of them \cite{Zhan2020AFF, Julian2020EfficientAF, Kalashnikov2021MTOptCM, Kumar2022PreTrainingFR}. Our work is most similar to Zhan et al.~\cite{Zhan2020AFF} in problem setting (RL with demonstrations) and experimental setup (robotic manipulation tasks in the real world). However, we focus on the unique challenges and opportunities of MBRL for real-world robot learning. Wu et al.~\cite{wu2023daydreamer} study real world MBRL, but consider simpler tasks with limited variation and do not leaverage demonstrations.

\section{Discussion}

In this work, we tackled the challenge of learning manipulation skills in the real world from only proprioceptive and visual feedback with sparse rewards. We developed MoDem-V2, a real-world ready adaptation of MoDem, by proposing to initially center rollouts around the BC policy, gradually increase the proportion of actions chosen by the learned world model, and implement uncertainty aware planning with actor-critic ensembles. We evaluated the sample-efficiency and safety of MoDem-V2 against strong baselines in simulation and found that it maintained the high sample-efficiency of MoDem while exhibiting significantly safer behavior through lower contact force exertion. We found that MoDem-V2 enabled a real, physical robot to learn a variety of manipulation skills, such as pushing, picking, and in-hand manipulation, from an hour or less worth of interaction data.


\textbf{Limitations}: One limitation of our work is that it requires a small number of demonstrations, which may not always be available or easy to collect. Also, this work assumes that the environment can be reset to a narrow set of starting states, which may not always be the case in the real-world. In future work, we hope to explore the reuse of our learned world model across changes in manipulated object and task goal.


\clearpage
\bibliography{bibliography}
\bibliographystyle{IEEEtran}


\clearpage
\section{Appendix}
\label{sec:appendix}
\subsection{MoDem Training Objective}

During online interaction, MoDem maintains a replay buffer $\mathcal{B}$ with trajectories, and the world model learns to minimize the following objective on length $h$ subtrajectories sampled from the replay buffer:

\begin{equation}
    \label{eq:tdmpc-loss}
    \mathcal{L}\left(\theta\right) \doteq \mathop{\mathbb{E}}_{\left(\mathbf{s}, \mathbf{a}, r, \mathbf{s}'\right)_{0:h} \sim \mathcal{B}} \left[ \sum_{t=0}^{h} \lambda^{t} \left( \mathcal{L}_{EM} + \mathcal{L}_{RE} + \mathcal{L}_{TD} \right)  \right]
\end{equation}

\begin{fleqn}
\begin{equation}
     \mathcal{L}_{EM} \doteq \mathcolor{black}{\|\ \mathbf{z}_{t}' - \operatorname{sg}(h_{\phi}(\mathbf{s}_{t}')) \|^{2}_{2}} {\color{CadetBlue}`\vartriangleleft\emph{Embedding Prediction}}
\end{equation}
\end{fleqn}
\begin{fleqn}
\begin{equation}
    \mathcal{L}_{RE} \doteq \mathcolor{black}{\|\hat{r}_{t} - r_{t}\|^{2}_{2}} {\color{CadetBlue}~~~~~~~~~~~~~~\vartriangleleft\emph{Reward Prediction}}
\end{equation}
\end{fleqn}
\begin{fleqn}
\begin{equation}
    \mathcal{L}_{TD} \doteq \mathcolor{black}{\|\hat{q}_{t} - q_{t}\|^{2}_{2}} {\color{CadetBlue}~~~~~~~~~~~~~~\vartriangleleft\emph{TD-learning}}
\end{equation}
\end{fleqn}

where $\operatorname{sg}$ is the \texttt{stop-grad} operator, $\phi$ is an exponentially moving average of $\theta$, $(\mathbf{z}_{t}',\hat{r}_{t},\hat{q}_{t})$ are as defined in Equation \ref{eq:tdmpc-components}, $q_{t} \doteq r_{t} + Q_{\phi}(\mathbf{z}_{t}', \pi_{\theta}(\mathbf{z}_{t}'))$ is the TD-target, and $\lambda \in (0,1]$ is a constant coefficient that assigns larger weight to temporally close time steps. Similarly, $\pi_{\theta}$ learns to maximize the objective $\mathcal{L}_{\pi}(\theta) \doteq \mathbb{E}_{\mathbf{s}_{0:h} \sim \mathcal{B}} \left[ \sum_{t=0}^{h} \lambda^{t} Q_{\theta}(\mathbf{z}_{t}, \pi_{\theta}(\mathbf{z}_{t})) \right],~\mathbf{z}_{t} = h_{\theta}(\mathbf{s}_{t})$ with gradients taken wrt. $\pi_{\theta}$ only.

\subsection{Additional Task Details}
\label{sec:add_task_details}
In this subsection, we provide additional details about the tasks explored in our experiments. For all experiments, the robot was controlled at a rate of 12.5 Hz, which corresponds to a time step length of 80 milliseconds. Each task used both proprioceptive and image inputs. Below we detail the action space, reset mechanism, demo collection, and reward specifications for each of the tasks.

\textbf{Planar Pushing:} The robot's action space is composed of the commanded absolute cartesian position and absolute yaw of the end effector. In simulation, the robot receives a reward of +1 for each timestep at which the object is within 5 centimeters of the goal position. For the real-world task, a computer vision module uses color thresholding in LUV color space to detect when the green object covers up the red goal area. A reward of +1 is assigned for each time step at which this occurs. At the end of each episode, the robot moves back to its starting configuration and the object is reset by two retractor reels that pull it back towards the starting position in order to reset it. We collect demonstrations for this task via teleoperation with an Oculus headset.

\textbf{Incline Pushing:} In this task, the robot's actions consist of the commanded cartesian position of the end effector. Here, rewards are specified and demos collected are in the same manner as the \textit{planar pushing} task. However this task differs in that it is reset purely by gravity; when the robot moves back to its starting configuration the block slides down the incline due to its own weight. This reset mechanism is less consistent in that it produces a much wider distribution over starting positions of the object relative to the \textit{planar pushing} task, resulting in a higher difficulty.

\textbf{Bin Picking:} Here, the robot's actions consists of the commanded absolute cartesian position and absolute yaw of the end effector, as well as commanding the gripper to either fully open or fully close. In simulation, a reward of +1 is assigned to each time step for which the object is within 7.5 centimeters away from a goal position above the table. To detect success at the end of an episode in the real-world, the robot moves out of the way in order to take a picture of the potentially empty bin, opens its gripper above the bin (allowing any object in its grasp to fall), and then takes another image of the bin. These two images are then subtracted and the result is thresholded to determine if the object was in the robot's grasp. If the episode was a success, we iterate through each timestep \textit{backwards in time} until we find a timestep at which the robot's gripper was less than 10 centimeters above the table. Each timestep later in time than this timestep is assigned a reward of +1. Note here that simply using gripper width to detect success would result in false positives because the robot's fingertips are somewhat flexible, as well as this particular gripper has two passive degrees of freedom. We used a hand-coded policy not only to collect demonstrations for this task, but also as a reset mechanism to ensure that the object was not pressed up against one of the sides of the bin at the beginning of an episode.

\textbf{In-Hand Reorientation:}  The robot's action space for this task includes the commanded absolute cartesian pose and absolute yaw of the end effector, and the absolute position of each of the 10 joints of the hand. The simulated environment assigns a reward of +1 to each time step at which the object has a roll and pitch less than 0.03 radians (i.e. an orientation that corresponds to the bottle being upright). In the real-world task we use the top-down depth camera to detect if the bottle has successfully been uprighted at the end of the episode. Given a successful episode, we iterate \textit{backwards through time} until we find a time step at which the hand was less than 10 centimeters above the table. Any time step after this time step will receive a reward of +1 if all three of the hand's metacarpal joints have angles less than a certain threshold corresponding to the hand being at least partially open and no longer grasping the bottle. Here the robot arm resets the task by knocking over the bottle if the episode succeeded and then moving it towards the center of the bin if necessary. We used a hand-coded policy to collect demonstrations for this task. The high-level strategy that the robot uses to achieve the task is to initially grasp the object near the bottle cap with its pinky and thumb fingers. Once it has lifted the object, it must strike a balance between applying sufficient force so as to not drop the object but also not too much force so that the bottle can pivot around the contact axis as the index finger pushes down on the bottle.

\subsection{Ineffectiveness of Torque Penalization}

Our results suggest that simply adding a reward penalty for exerted torque is ineffective at reducing safety violations and results in a safety profile very similar to the original MoDem, as shown in \autoref{fig:torque_penalty_series} below. Here we swept over 3 orders of magnitude for an additive reward term that penalizes high torques. This reiterates the pivotal role played by our three proposed modifications in enabling successful learning on physical robots.

\begin{figure}[H]
  \includegraphics[width=\linewidth]{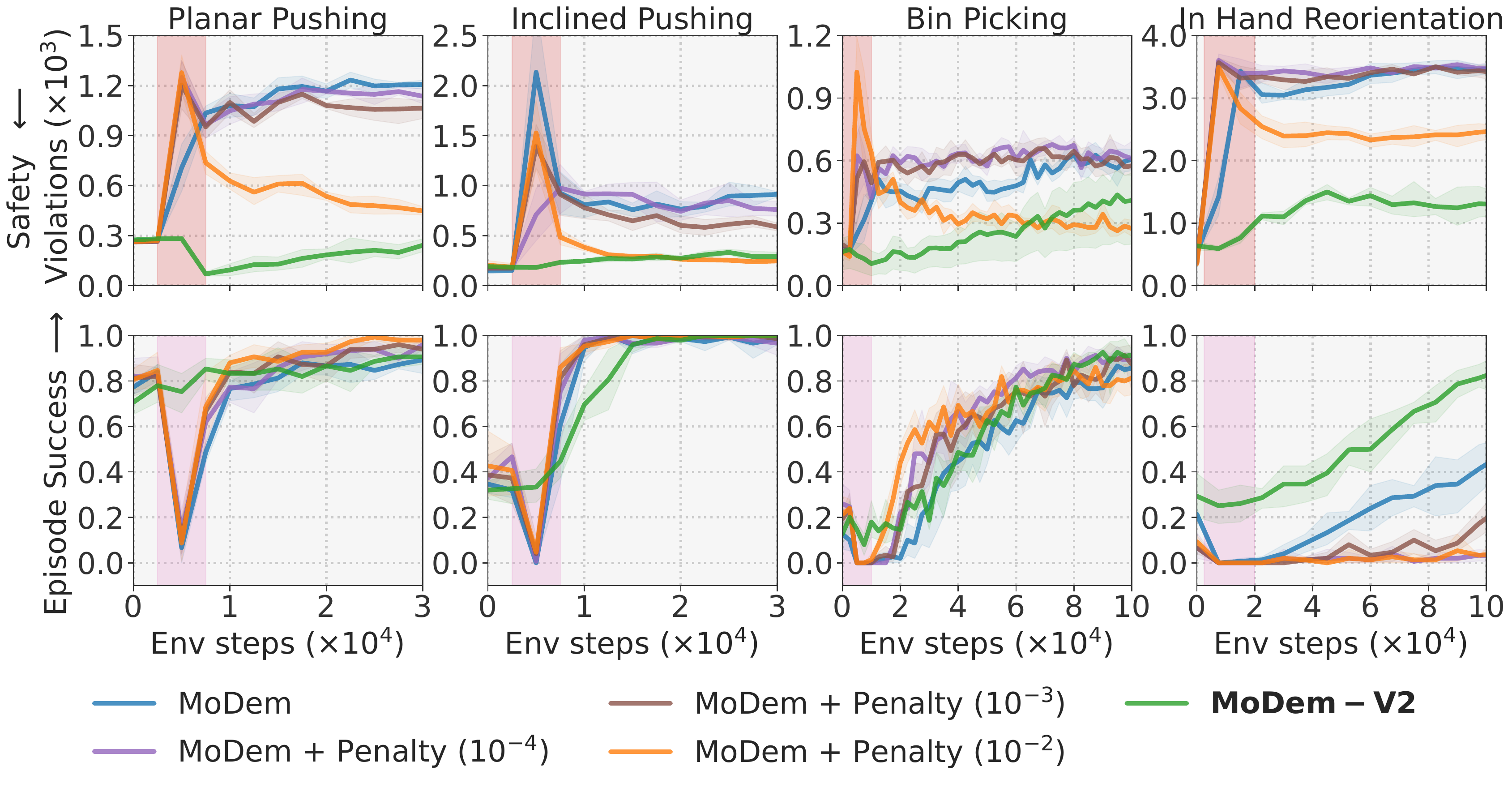}
  \caption{The number of safety violations (top row) and success rate (bottom row) for each of the four manipulation tasks in simulation. Lower is better for safety violations (top row) while higher is better for episode success (bottom row). On top of MoDem-v1, we add an additional term to the reward that penalizes the use of high torques (additive L2 penalty). While penalization can help reduce torque in later stages of training, it does not prevent a large spike in exerted torque at the beginning of online interaction as indicated by the region shaded in red. This also correlates with a large drop in performance as indicated by the region shaded in pink. Furthermore, in the hardest in-hand reorientation task, the use of additive penalty completely stagnates learning and results in an unsuccessful policy.}
  \label{fig:torque_penalty_series}
\end{figure}

\subsection{Task Trajectories and Learning Curves}
\autoref{fig:train_plots} below shows the learning curves of our MoDem-V2 agent on real world tasks.

\begin{figure}[H]
  \includegraphics[width=\linewidth]{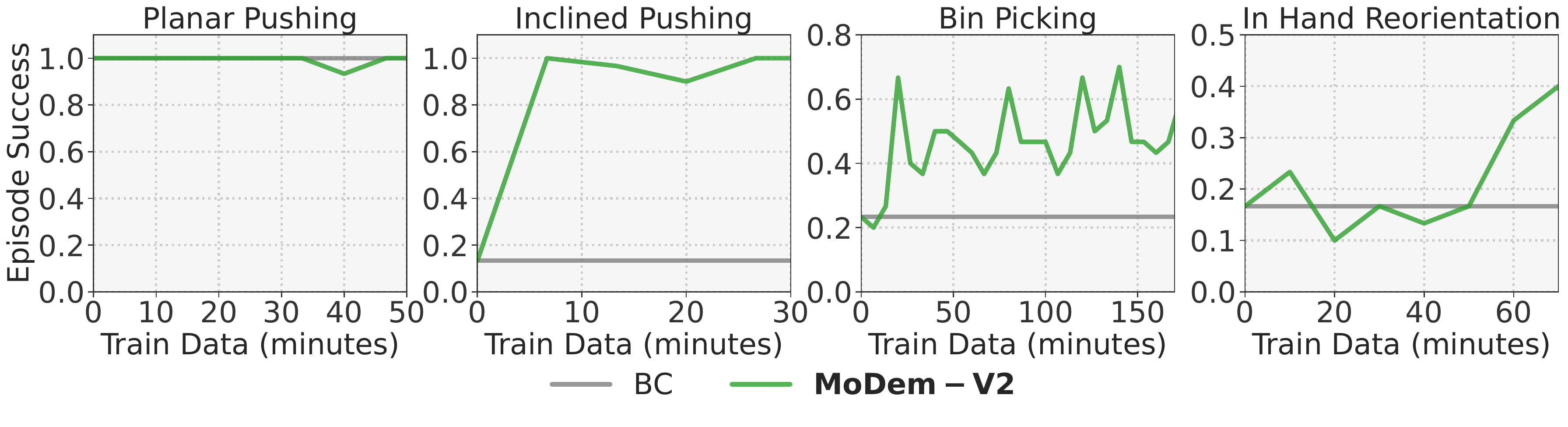}
  \caption{MoDem-V2 training performance on real-world manipulation tasks.}
  \label{fig:train_plots}
\end{figure}

\newpage

\subsection{MoDem-V2 Hyperparameters}


\begin{table} [h]
\centering
\vspace{-0.225in}
\caption{\textbf{MoDem hyperparameters.} We list all relevant hyperparameters for MoDem-V2 below. \colorbox{citecolor!15}{Highlighted} rows are unique to MoDem-V2, while the others are inherited from TD-MPC and MoDem but included for completeness.}
\label{tab:ours-hparams}
\vspace{0.05in}
\centering
\begin{tabular}{@{}ll@{}}
\toprule
Hyperparameter                                                                   & Value                                                                             \\ \midrule
Discount factor ($\gamma$)                                                         & 0.95                                                                              \\
Image resolution                                                          & $224\times224$ \\
Frame stack                                                               & $2$ \\
Data augmentation                                                         & $\pm10$ pixel image shifts \\
Action repeat                                                             & $1$ \\
\begin{tabular}[c]{@{}l@{}}Seed steps\\~ \end{tabular}                                                                    & \begin{tabular}[c]{@{}l@{}}7500 (In-hand reorientation)\\5000 (otherwise)\end{tabular}\\                                                             
 Pretraining objective               &  Behavior cloning \\
 Seeding policy                      &  Behavior cloning \\
 Number of demos                                          &  $10$                                                                          \\
 Demo sampling ratio                                          &  $75\% \rightarrow 25\%$ (100K steps)                                                                          \\
Replay buffer size                                                        & Unlimited                                                                                      \\
Sampling technique                                                        & PER ($\alpha=0.6, \beta=0.4$)                                                                                      \\
Planning horizon ($H$)                                                         & $5$                                                                              \\
\colorcellh \begin{tabular}[c]{@{}l@{}}MPC action probability ($\alpha$)\\~ \end{tabular}                                                                    & \colorcellh\begin{tabular}[c]{@{}l@{}}$0.0\rightarrow 1.0$ (25k steps for sim \\~~~~~~~~~~~~~~~incline pushing) \\$0.0\rightarrow 1.0$ (else 100k steps) \end{tabular}\\   
\colorcellh Critic mean weight ($w_1$) & \colorcellh 1.0 \\
\colorcellh Critic std dev weight ($w_2$) & \colorcellh -10.0 \\

Initial parameters ($\mu^{0}, \sigma^{0}$)                                 & $(0, 2)$                                                                          \\
Population size                                                            & $510$                                                                          \\
Elite fraction                                                             & $64$                                                                          \\
Iterations & $1$ \\

Policy fraction                                                            & $5\%$                                                                          \\
Number of particles                                                        & $1$                                                                          \\
Momentum coefficient                                                       & $0.1$                                                                          \\
Temperature ($\tau$)                                                 & $0.5$                                                                         \\
MLP hidden size                                                                & $512$                                                                          \\
MLP activation                                                                 & ELU
                                        \\
Latent dimension & $50$ \\
\colorcellh Ensemble size & \colorcellh $5$ \\
Learning rate                                                       & 3e-4\\ 
Optimizer ($\theta$)                                                & Adam ($\beta_1=0.9, \beta_2=0.999$)                                                       \\
Temporal coefficient ($\lambda$)                                              & $0.5$                                                                          \\
Reward loss coefficient ($c_{1}$)                                              & $0.5$                                                                          \\
Value loss coefficient ($c_{2}$)                                              & $0.1$                                                                          \\
Consistency loss coefficient ($c_{3}$)                                        & $2$                                                                          \\
Exploration schedule ($\epsilon$)                                             & $0.1\rightarrow 0.05$ (25k steps)                                                                          \\
\begin{tabular}[c]{@{}l@{}}Batch size\\~ \end{tabular}                                                                    & \begin{tabular}[c]{@{}l@{}}256 (All sim, and real-world \\~~~~~~~in-hand reorientation) \\64 (else due to mem constraints)\end{tabular}\\  
Momentum coefficient ($\zeta$)                                                 & $0.99$                                                                         \\
Steps per gradient update                                                     & $1$                                                                          \\
$\Bar{\theta}$ update frequency                                                  & 2                                                                 \\ \bottomrule
\end{tabular}%
\vspace{-0.15in}
\end{table}







\end{document}